\documentclass[journal]{IEEEtran}
\usepackage{amsfonts}
\usepackage{caption}
\usepackage{amsmath}
\usepackage{multirow}
\usepackage{multicol}
\usepackage{booktabs}
\usepackage[hyphens]{url}
\usepackage{breakurl}
\usepackage{float}
\usepackage{amsmath}
\usepackage{setspace}
\usepackage{lineno}
\usepackage{cite}
\ifCLASSINFOpdf
   \usepackage[pdftex]{graphicx}
   \graphicspath{{./}}
   \DeclareGraphicsExtensions{.pdf,.jpeg,.png}
\else
\fi
\usepackage{array}
\usepackage{url}

\begin{document}
%
\title{ResiDualGAN: Resize-Residual DualGAN for Cross-Domain Remote Sensing Images Semantic Segmentation}
%
%
%

\author{Yang Zhao,
        Peng Guo,
        Zihao Sun, 
        Xiuwan Chen,
        Han Gao,
\thanks{Y. Zhao, P. Guo, Z. Sun, X. Chen, and H. Gao are with the Institute of Remote Sensing and Geographic Information System, Peking University, Beijing 100871, China. E-mail: zy\_@pku.edu.cn; peng\_guo@pku.edu.cn; sunzihao211@gmail.com; xwchen@pku.edu.cn; hgao@pku.edu.cn;.}
\thanks{Corresponding author:Han Gao (hgao@pku.edu.cn).}
\thanks{© 2022 IEEE. Personal use of this material is permitted. Permission from IEEE must be obtained for all other uses, including reprinting/republishing this material for advertising or promotional purposes, collecting new collected works for resale or redistribution to servers or lists, or reuse of any copyrighted component of this work in other works.}
}

\maketitle

\begin{abstract}
The performance of a semantic segmentation model for remote sensing (RS) images pretrained on an annotated dataset would greatly decrease when testing on another unannotated dataset because of the domain gap. Adversarial generative methods, e.g., DualGAN, are utilized for unpaired image-to-image translation to minimize the pixel-level domain gap, which is one of the common approaches for unsupervised domain adaptation (UDA). However, the existing image translation methods are facing two problems when performing RS images translation: 1) ignoring the scale discrepancy between two RS datasets which greatly affects the accuracy performance of scale-invariant objects, 2) ignoring the characteristic of real-to-real translation of RS images which brings an unstable factor for the training of the models. In this paper, ResiDualGAN is proposed for RS images translation, where an in-network resizer module is used for addressing the scale discrepancy of RS datasets, and a residual connection is used for strengthening the stability of real-to-real images translation and improving the performance in cross-domain semantic segmentation tasks. Combined with an output space adaptation method, the proposed method greatly improves the accuracy performance on common benchmarks, which demonstrates the superiority and reliability of ResiDuanGAN. At the end of the paper, a thorough discussion is also conducted to give a reasonable explanation for the improvement of ResiDualGAN. Our source code is available at \url{https://github.com/miemieyanga/ResiDualGAN-DRDG}. 
\end{abstract}


%
\IEEEpeerreviewmaketitle

\section{Introduction}
\IEEEPARstart{W}{ith} the development of unmanned aerial vehicle (UAV) photography and remote sensing (RS) technology, the amount of very-high-resolution (VHR) RS images has increased explosively\cite{gao_classification_2021}. Semantic segmentation is a vital application area for VHR RS images, which gives a pixel-level ground classes classification for every image. Followed by AlexNet\cite{krizhevsky_imagenet_2012}, convolutional neural networks (CNN) based methods, learning from part annotated data and predicting data in the same dataset, show great advantages compared with traditional methods when performing semantic segmentation tasks\cite{long_fully_2015,ronneberger_u-net_2015,chen_encoder-decoder_2018}, which also bring a giant promotion of semantic segmentation for VHR RS images\cite{yuan_review_2021}. \par
Nevertheless, though great success has been made, disadvantages of CNN-based methods are also obvious, such as laborious annotation, poor generalization, and so on\cite{guo_review_2018}. Worse still, the characteristic of laborious annotation and poor generalization may be magnified in the RS field\cite{zhang_curriculum-style_2021}. With more and more RS satellites being launched and UAVs being widely used, RS images are produced with various sensor types in different heights, angles, geographical regions, and at different dates or times in one day\cite{li_learning_2021}. As a result, RS images always show a domain discrepancy mutually. When a well-trained CNN module is applied to a different domain, the performance is most likely to decline due to the gap between the two domains\cite{yao_weakly-supervised_2021}. However, annotation is a laborious and time-wasting job that is not likely to be obsessed by every RS dataset\cite{zhang_curriculum-style_2021}. Hence, how to minimize this kind of discrepancy between domains and fully utilize these non-annotated data is now a hotspot issue in the RS field.\par

To this end, unsupervised domain adaptation (UDA) has been proposed in the computer vision (CV) field to align the discrepancy between the source and the target domain. Approaches of UDA can be roughly divided into four categories: adversarial generative method\cite{bousmalis_unsupervised_2017,shrivastava_learning_2017,yan_pixel-level_2021}, adversarial discriminative method\cite{ganin_unsupervised_2015,tsai_learning_2018,wang_classes_2020}, semi-supervised and self-learning method\cite{zou2018unsupervised,tranheden_dacs_2021}, and others\cite{hoyer2022daformer}. In this paper, we mainly focus on the former two categories. The adversarial generative method minimizes the discrepancy between two domains in the pixel level, which makes the images of two domains resemble each other at a low level. Inspired by the generative adversarial networks(GANs)\cite{goodfellow_generative_2014}, CyCADA\cite{bousmalis_unsupervised_2017} uses CycleGAN\cite{zhu_unpaired_2017} to diminish the pixel level discrepancy, outperforming others methods at that time. The adversarial discriminative method is another common approach for UDA. The adversarial discriminative method minimizes the domain gap in the feature and output level. Ganin's\cite{ganin_unsupervised_2015} work tries to align feature space distribution via a domain classifier. AdaptSegNet\cite{tsai_learning_2018} significantly improves the performace by output space discriminating. FADA\cite{wang_classes_2020} propose a fine-grained adversarial learning strategy for feature-level alignment. Recently, self-learning and Transformer-based methods show superority in the CV field. CBST\cite{zou2018unsupervised} propose an iterative self-training procedure alternatively generating pseudo labels on target data and re-training the model with these labels. DAformer\cite{hoyer2022daformer} introduces Transformer\cite{vaswani2017attention} to the UDA problem. \par

In the field of RS, some attempts have been made \cite{benjdira_unsupervised_2019,ji_generative_2021,li_learning_2021,yao_weakly-supervised_2021,zhang_curriculum-style_2021,shi_end--end_2021,shi_rotation_2021,tasar2020colormapgan,lubin,wittich2021appearance}. 
Benjdira's\cite{benjdira_unsupervised_2019} work first introduces CycleGAN into the cross-domain semantic segmentation of RS images, validating the feasibility of using image translation to minimize the domain gap between two RS datasets. FSDAN\cite{ji_generative_2021} and MUCSS\cite{li_learning_2021} follow the routine of Benjdira's work. FSDAN extends the pixel-level adaptation to both the feature level and output level. MUCSS utilizes the self-training strategy to further improve performance. Except for generative methods,  Bo's\cite{zhang_curriculum-style_2021} work explores curriculum learning to accomplish the feature alignment process from the locally semantic to globally structural feature discrepancies. Lubin's\cite{lubin} work incorporates comparative learning with the framework of UDA and achieves better accuracy. For most of the recent work, the generative methods tend to be gradually abandoned because of their instability and deficiency. However, in this paper, we reconsider the merits of the generative models. By simply modifying the structure of the existing generative model, the proposed generative method surpasses all the other methods. 

Utilizing the GANs-based UDA methods and self-training strategies, the performance of cross-domain semantic segmentation of VHR RS images has greatly improved. However, compared with the CV field, RS images have some unique features which should be processed specifically, while most of the architectures of networks used for RS image-to-image translation are directly carried from the CV field, such as CycleGAN and DualGAN\cite{yi_dualgan_2017}, which may bring about the following problems:\par
First, ignoring the scale discrepancy of RS images datasets. RS images in a single domain were taken at a fixed height using a fixed camera focal length, and the object distance for any objects in RS images is a constant number, leading to a scale discrepancy between two RS images datasets. As a comparison, images in the CV field were mostly taken from a portable camera or an in-car camera, objects may be close to the camera, however, may also be away from the camera, which provides a varied scale for all kinds of objects. Consequently, traditional image translation methods carried from the CV field are not suitable for the translation of RS images which may cause the accuracy decrease for some specific classes. Some previous works attempt to utilize the images resize process as a pre-processing before feeding them forward into networks\cite{ji_generative_2021,wittich2021appearance}. However, this  pre-processing may lead to information loss of images, resulting in a performance decline of the semantic segmentation model.  \par

\begin{figure}
    \centering
    \setlength{\abovecaptionskip}{0pt}
    \setlength{\belowcaptionskip}{0pt}
    \includegraphics[width=.43\textwidth]{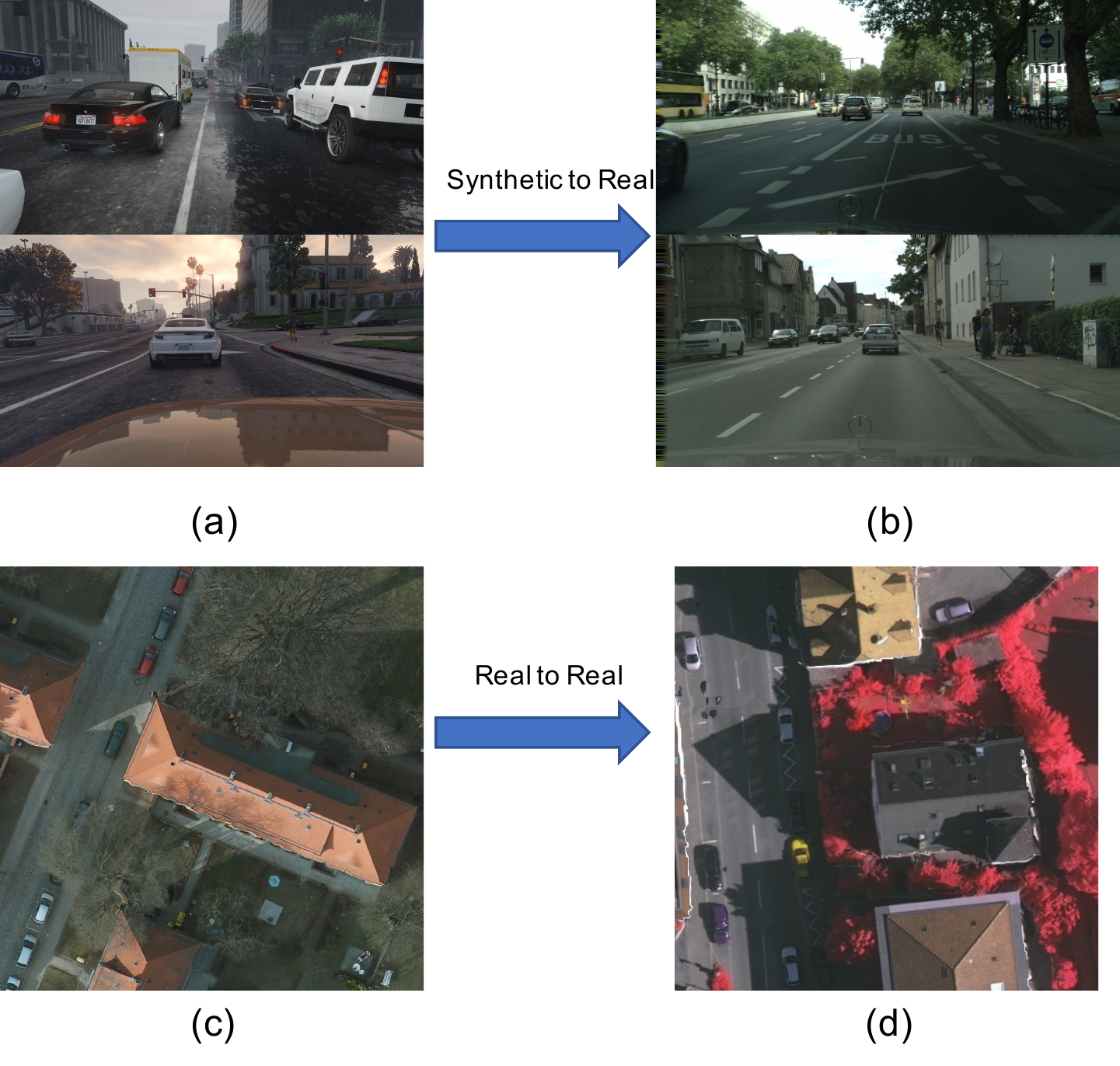}
    \caption{Synthetic-to-real and real-to-real translation. (a) GTA5. (b) Cityscapes. (c) PotsdamRGB. (d) Vaihingen. Where GTA5 is a commonly used computer-synthesized dataset, the typical task of UDA in the computer vision field is to train a model on GTA5 and deploy it on the natural world scene like Cityscapes. As a comparison, the mission of UDA of RS images is always the real-world scene (e.g. Potsdam) to another real-world scene(e.g. Vaihingen). }
    \label{fig:dataset}
\end{figure}

Second, underutilizing the feature of real-to-real translation of RS images (Fig. \ref{fig:dataset}). CycleGAN or DualGAN and many other image-to-image translation networks were initially designed to carry out not only a real-to-real translation, but also a synthetic-to-real translation, such as photos to paints, or game to the real world, while the RS images translation is always real to real, both two sides are real-world images that are geographical significant. Synthetic-to-real translation is likely to bring a structure information change to a certain object because of the architecture of generators, which disturbs the task of segmentation. In addition, the gap of the marginal distribution of synthetic and real is larger than real and real, such as generating a Cityscapes\cite{cordts_cityscapes_2016} stylized image from GTA5\cite{richter_playing_2016} where the generator is too overloaded to generate a new image, which could be largely avoided in the RS images translation because both sides are from the real world.\par

Aimed at the two problems aforementioned, this paper proposed a new architecture of GANs based on DualGAN, named ResiDualGAN, for RS images domain  translation and cross-domain semantic segmentation. ResiDualGAN resolves the first problem proposed above by using an in-network resizer module to fully utilize the scale information of RS images. And a residual connection that transfers the function of the generator from generating new images to generating residual items is used which addresses the second problem. By combining with other methods simply, our proposed method reaches the state-of-the-art accuracy performance in the common dataset. The main contributions of the paper can be summarized as follows:
\begin{enumerate}
\item A new architecture of GANs, ResiDualGAN, is implemented based on DualGAN to carry out unpaired RS images cross-domain translation and cross-domain semantic segmentation tasks, in which an in-network resizer module and a residual architecture are used to fully utilize unique features of RS images compared with images used in the CV field. The experiment results show the superiority and stability of the ResiDualGAN. 
\item To the best of our knowledge, the proposed method reaches state-of-the-art performance when carrying out a cross-domain semantic segmentation task between two open-source datasets: Potsdam and Vaihingen\cite{potsdam_vaihingen}. The mIoU and F1-score are 55.83\% and 68.04\% respectively when carrying out segmentation task from PotsdamIRRG to Vaihingen, promoting 7.33\% and 4.94\% compared with state-of-the-art methodology. 
\item On the foundation of thorough experiments and analyses, this paper attempts to explain the reason why such great improvement could be achieved by implementing this kind of simple modification, which should be specially noted when a method from the CV field is applied to RS images processing.
\end{enumerate}


\section{Method}
\begin{figure*}
    \centering
    \setlength{\abovecaptionskip}{0pt}
    \setlength{\belowcaptionskip}{0pt}
    \includegraphics[width=0.90\textwidth, height=0.60\textheight]{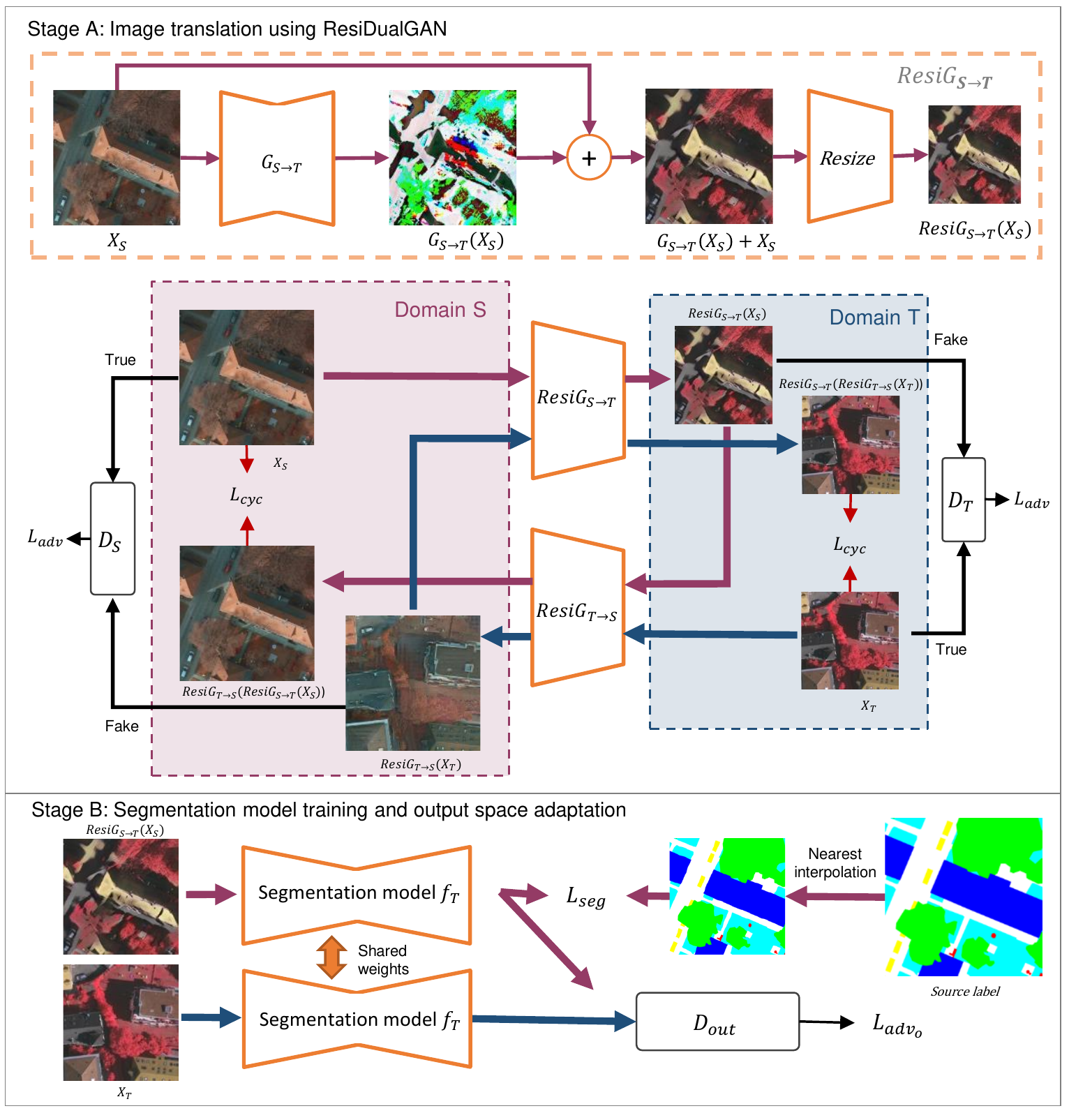}
    \caption{Overview of the proposed method.}
    \label{fig:overview}
\end{figure*}
To describe the proposed methodology more specifically, some notions used in this paper should be defined first. Let $X_S\in\mathbb{R}^{H_S{\times}W_S{\times}B}$ be images from the source domain $S$ with a resolution of $r_S$, where $B$ is the number of channels. And let $X_T\in\mathbb{R}^{H_T{\times}W_T{\times}B}$ be images from the target domain $T$, with resolution $r_T$. $Y_S\in\mathbb{Z}^{H_S{\times}W_S{\times}C}$ are the labels of $X_S$, where $C$ is the number of classes, while no labels for $X_T$. For simplifying the problem, we propose the following formulation. The size and the resolution of images should conform to this formulation to diminish the effect of the scale factor:
\begin{equation}
\frac{H_S}{H_T}=\frac{W_S}{W_T}=\frac{r_T}{r_S}\label{formu:1}
\end{equation}
\par
The objective of the proposed methodology is, for any given images $X_T$ in the target domain, we want to find a semantic segmentation model $f_T:\ X_T\!\rightarrow\!Y_T\in\mathbb{Z}^{H_T{\times}W_T{\times}C}$, which is expected to generate prediction labels $Y_T$ for $X_T$. The overview of the proposed method is shown in Fig. \ref{fig:overview}, where two separated stages are implemented. Stage A is used to carry out an unpaired images style transfer from $S$ to $T$ with ResiDualGAN, which is proposed by this paper. Stage B trains a semantic segmentation model $f_T$ by utilizing the style transferred images obtained from stage A with their respective labels $Y_S$. Additionally, an output space adaptation (OSA) proposed by\cite{tsai_learning_2018} is applied during stage B, which is an more effective way to improve the performance of cross-domain semantic segmentation models compared with the feature level adaptation used in many RS research\cite{ji_generative_2021}.

\subsection{Stage A: Image Translation Using ResiDualGAN}
\subsubsection{Overall}
The objective of Stage A is to translate $X_S$ to the style of $X_T$. Inspired by DualGAN\cite{yi_dualgan_2017}, which is proven to be the optimal choice for VHR RS images translation\cite{li_learning_2021}, ResiDualGAN is proposed for VHR RS images translation, which consists of two major components: ResiGenerators and discriminators. ResiGenerator is exploited to generate a style transferred image while discriminator is designed to discern whether an image is generated by ResiGenerator or not. ResiDualGAN consists of two ResiGenerators: $ResiG_{S\!\rightarrow\!T}$, $ResiG_{T\!\rightarrow\!S}$, and two discriminators: $D_S$, $D_T$. $ResiG_{S\!\rightarrow\!T}$ is used for translating images from $S$ to $T$, while $ResiG_{T\!\rightarrow\!S}$ is $T$ to $S$. $D_S$ performs the task of discerning whether an image is from S or being generated by $ResiG_{T\!\rightarrow\!S}$, while $D_T$ discerns whether an image is from $T$ or being generated by $ResiG_{S\!\rightarrow\!T}$. 
\subsubsection{ResiGenerator}
The architecture of $ResiG_{S\!\rightarrow\!T}:$\\$\mathbb{R}^{H_S{\times}W_S{\times}B}\!\rightarrow\!\mathbb{R}^{H_T{\times}W_T{\times}B}$ is illustrated in Fig. \ref{fig:overview}, containing a generator $G_{S\!\rightarrow\!T}$ based on U-Net\cite{ronneberger_u-net_2015} and a resizer module $Resize_{S\!\rightarrow\!T}$.$G_{S\!\rightarrow\!T}:\mathbb{R}^{H_S{\times}W_S{\times}B}\!\rightarrow\!\mathbb{R}^{H_S{\times}W_S{\times}B}$ generates a residual item for its input. $Resize_{S\!\rightarrow\!T}:$ $\mathbb{R}^{H_S{\times}W_S{\times}B}$\\$\!\rightarrow\!\mathbb{R}^{H_T{\times}W_T{\times}B}$ is a resizing function that resizes images of the source domain to the size of the target domain, implemented as a network or an interpolation function. Based on ablation experiments in the follows of the paper, a bilinear interpolation function is used as the resizing function eventually. For the source domain images $X_S$, we want to get their corresponding target-stylized images $X_{S\!\rightarrow\!T}$ using $ResiG_{S\!\rightarrow\!T}$, which can be written as the following formulation:
\begin{equation}
\begin{split}
        X_{S\!\rightarrow\!T}=&ResiG_{S\!\rightarrow\!T}\left(X_S\right)\\=&Resize_{S\!\rightarrow\!T}\left(G_{S\!\rightarrow\!T}\left(X_S\right)+k\times X_S\right)\ 
    \label{formu:2}
\end{split}
\end{equation}
Where $k$ is the hyperparameter. All of the random noises are simplified for facilitating expression. According to the hypothesis in Equation \eqref{formu:1}, after the resizing operation, the resolution of $X_{S\!\rightarrow\!T}$ should be the same as $X_T$.
The architecture of $ResiG_{T\!\rightarrow\!S}$ just resembles with $ResiG_{S\!\rightarrow\!T}$, where the only remaining difference is the resizer module $Resize_{T\!\rightarrow\!S}:\mathbb{R}^{H_T{\times}W_T{\times}B}\!\rightarrow\!\mathbb{R}^{H_S{\times}W_S{\times}B}$. $Resize_{T\!\rightarrow\!S}$ is a reverse procedure of $Resize_{S\!\rightarrow\!T}$, which resizes an image with the size of $H_T{\times}W_T{\times}B$ to the size of $H_S{\times}W_S{\times}B$. Consequently, the size and resolution of $ResiG_{T\!\rightarrow\!S}\left(X_{S\!\rightarrow\!T}\right)$ should be the same as $X_S$.
\subsubsection{Adversarial Loss}
The ResiGenerator attempts to generate images to deceive the discriminator, while discriminator attempts to distinguish whether images generated by ResiGenerator or not, resulting in an adversarial loss $L_{adv}$. Analogous with DualGAN, a Wasserstein-GAN (WGAN) loss proposed by\cite{arjovsky_wasserstein_2017} is used to measure the adversarial loss. WGAN resolves the problem that the distance may be equaled with 0 when there is no intersection of two data distributions by utilizing the Earth-Mover (EM) distance as a measure for two data distributions, which avoids the vanishing gradients problem of networks. Differentiate with traditional GANs\cite{goodfellow_generative_2014} using a sigmoid output as an finally output for discriminator, discriminators of DualGAN remove sigmoid as a final layer, which is also used by this paper. 
Moreover, a gradient penalty proposed by\cite{gulrajani_improved_2017} is used to stabilize the training procedure and avoid the weights clipping operation in WGAN which is in order to enforce the Lipschitz constraint in WGAN. The gradient penalty is not written in the formulations for the purpose of simplification. 
Above all, the adversarial loss $L_{adv}$ of source domain $S$ to target domain $T$ can be written as follows:
\begin{equation}
\begin{split}
    L_{adv}^{S\rightarrow T}=
    &\mathbb{E}_{x_T\!{\sim}\!X_T}(D_T(x_T))-\\
    &\mathbb{E}_{x_S\!{\sim}\!X_S}(D_T({ResiG}_{S\!\rightarrow\!T}(x_S)))
    \label{formu:3}
\end{split}    
\end{equation}
\subsubsection{Reconstruction Loss}
$ResiG_{T\!\rightarrow\!S}$ reconstructs $X_{S\!\rightarrow\!T}$ to the style of the source domain. Ideally, $ResiG_{T\!\rightarrow\!S}\left(X_{S\!\rightarrow\!T}\right)$ should be entirely same as $X_S$. However, loss always exits. Considering both two sides of ResiDualGAN, the reconstruction loss $L_{cyc}$ can be measured by an L1 penalty as follows:
\begin{equation}
\begin{split}
    &L_{cyc}=\mathbb{E}_{x_S\!{\sim}\!X_S\!}(\Vert ResiG_{T\!\rightarrow\!S}(ResiG_{S\!\rightarrow\!T}(x_S)))-x_S \Vert_1)\\
    &+\mathbb{E}_{x_T\!{\sim}\!X_T\!}(\Vert ResiG_{S\!\rightarrow\!T}(ResiG_{T\!\rightarrow\!S}(x_T))-x_T \Vert_1)
    \label{formu:4}
\end{split}    
\end{equation}
\subsubsection{Total Loss}
Considering adversarial loss and reconstruction loss together, the total loss for ResiDualGAN can be written:
\begin{equation}
\begin{split}
    L_{ResiG}=
    \lambda_{cyc}L_{cyc}
    +\lambda_{adv}(L_{adv}^{S\rightarrow T}+L_{adv}^{T\rightarrow S})
    \label{formu:5}
\end{split}    
\end{equation}
Where $\lambda_{cyc}$ and $\lambda_{adv}$ are hyperparameters corresponding to the reconstruction loss and the adversarial loss. At the training of ResiDualGAN, the discriminators attempts to maximize $ L_{ResiG}$ while the ResiGenerator attempts to minimize, which can be written as the following min-max criterion:
\begin{equation}
\begin{split}
    \mathop{max}\limits_{\substack{D_S\\D_T}} \mathop{min}\limits_{\substack{ResiG_{S\rightarrow T}\\ ResiG_{T\rightarrow S}}} L_{ResiG}
    \label{formu:6}
\end{split}    
\end{equation}
After the training of Stage A, a ResiGenerator $ResiG_{S\!\rightarrow\!T}$ can be obtained to perform the style translation task in Equation \eqref{formu:2} and $X_{S\!\rightarrow\!T}$ is generated for the next training in Stage B, which is uncoupled with the training of Stage A. 

\subsection{Stage B: Segmentation Model Training}
\subsubsection{Overall}
The objective of Stage B is to find the optimal model $f_T$ for semantic segmentation in the target domain (Fig. \ref{fig:overview}, Stage B). In Stage B, an OSA is performed where we can regard $f_T$ as a generator in traditional GANs, which generates softmax prediction outputs for both $X_{S\!\rightarrow\!T}$ and $X_T$. Meanwhile the function of discriminator $D_{out}$ in Stage B is to discern whether the output of $f_T$ is generated by $X_{S\!\rightarrow\!T}$ or $X_T$. OSA assumes that while images may be very different in appearance, their outputs are structured and share many similarities, such as spatial layout and local context. In this task, OSA minimizes the output space gap between $f_T(X_{S\!\rightarrow\!T})$ and $f_T(X_T)$ and significantly improves the segmentation accuracy on $X_T$.As well as the traditional GANs, at the beginning, we will train $D_{out}$ firstly and then $f_T$.
\subsubsection{Discriminator Training}
A binary cross-entropy loss $L_{out}$ as loss is used for $D_{out}$.
\begin{equation}
\begin{split}
    L_{out}=-\mathbb{E}_{x_T{\sim}X_T}(log(1-D_{out}(f_T(x_T))))\\
    -\mathbb{E}_{x_{S\!\rightarrow\!T}{\sim}X_{S\!\rightarrow\!T}}(log(D_{out}(f_T(x_{S\!\rightarrow\!T}))))
    \label{formu:7}
\end{split}    
\end{equation}
\subsubsection{Semantic Segmentation Training}
At first, A cross-entropy loss is used to train $f_T$. Because of the shape of $X_{S\!\rightarrow\!T}$ is $H_T{\times}W_T{\times}B$ while the shape of the label $Y_S$ is $H_S{\times}W_S{\times}C$, a nearest interpolation method is performed to resize $Y_S$ as $Y_{S\!\rightarrow\!T}\in\mathbb{Z}^{H_T{\times}W_T{\times}B}$. Then the segmentation loss is:
\begin{equation}
\begin{split}
    &L_{seg}=\\
    &-\mathbb{E}_{(x_{S\!\rightarrow\!T},y_{S\!\rightarrow\!T}){\sim}(X_{S\!\rightarrow\!T},Y_{S\!\rightarrow\!T})}(\sum_{C}^{c=1}{y_{S\!\rightarrow\!T}log(f_T(x_{S\!\rightarrow\!T}))})
    \label{formu:8}
\end{split}    
\end{equation}
Where $y_{S\!\rightarrow\!T}\in Y_{S\!\rightarrow\!T}$ is the nearest interpolation resizing of $y_S$, which is the label for $x_S\rightarrow T$. 
Next, we forward $X_T$ to $f_T$ and obtain an output $f_T\left(X_T\right)$, which attempts to fool the discriminator $D_{out}$, resulting in an adversarial loss $L_{adv_o}$:
\begin{equation}
\begin{split}
    L_{adv_o}=\mathbb{E}_{x_T{\sim}X_T}(log(1-D_{out}(f_T(x_T))))
    \label{formu:9}
\end{split}    
\end{equation}
\subsubsection{Total loss}
Considering all items in Stage B, the total loss for the semantic segmentation task can be written as:
\begin{equation}
\begin{split}
    L_{total}=\lambda_{seg}L_{seg}+\lambda_{adv_o}L_{adv_o}
    \label{formu:10}
\end{split}    
\end{equation}
Where $\lambda_{seg}$ and $L_{adv_o}$ are hyperparameters corresponding to the segmentation loss and the the OSA loss. At the training of Stage B, we optimize the following min-max criterion:
\begin{equation}
\begin{split}
    \mathop{max}\limits_{D_{out}} \mathop{min}\limits_{f_T} L_{total}
    \label{formu:min-max_L_total}
\end{split}    
\end{equation}
After the training of Stage B, a semantic segmentation model $f_T$ for target images $X_T$ can be obtained finally.
\subsection{Networks Settings}
\subsubsection{ResiGenerators}
A ResiGenerator consists of a generator and an in-network resizer module. The generator is implemented as an U-Net\cite{ronneberger_u-net_2015}, which is a fully convolutional network with skip connections between down-sampling and up-sampling layers. For down-sampling layers, we set the size of the convolving kernel as 4, padding as 1, and strides as 2. The channels of layers are \{64, 128, 256, 512, 512, 512, 512\}, where except for the first layer and the final layer, all layers are followed by an instance normalization\cite{ulyanov_instance_2016}, and a Leaky ReLU\cite{maas_rectifier_2013} with a negative slope of 0.2. For up-sampling layers, the same convolving kernel size, stride, and padding from down-sampling layers are utilized. All layers are composed of a transposed convolutional layer followed by an instance normalization and a ReLU\cite{nair_rectified_2010}. Dropout layers with a probability of 0.5 are exploited in all up-sampling and down-sampling layers with channels of more than 256.  
The resizer module could be implemented in many approaches. Based on the experimental results in section \ref{resizer}, we exploit the bilinear interpolation as the optimal implementation of the resizer module eventually.
\subsubsection{Discriminators}
Discriminators of ResiDualGAN are implemented as fully convolutional networks as well. The channels of layers are \{64, 128, 256, 512, 512, 1\}, where except for the last layer, all the layers are followed with a batch normalization\cite{ioffe_batch_2015} layer and a Leaky ReLU with a negative slope of 0.2. And the output of the last layer doesn’t pass through any activation functions due to the constraints of WGAN\cite{arjovsky_wasserstein_2017}.

\subsubsection{Output Space Discriminator}
The implementation is totally the same with \cite{tsai_learning_2018}, which is a fully convolutional network with the kernel of 4 × 4, the stride of 2, and channels of {64, 128, 256, 512, 1}. Except for the last layer, every layer is followed with Leaky ReLU with a negative slope of 0.2. And the output of the last layer is resized to the size of the input. 
\subsubsection{Segmentation Baseline}
DeeplabV3\cite{chen2017rethinking} is adopted as the baseline network for the semantic segmentation task of the proposed method. To accelerate the coverage procedure, the encoder of the baseline is replaced as ResNet-34\cite{he_deep_2016} which is pre-trained on ImageNet\cite{deng_imagenet_2009}.
\subsection{Training Settings}
All the models were implemented on PyTorch 1.8.1 and trained on NVIDIA A30 with 24GB RAM under ubuntu 18.04. The total time consumption is about 23 hours, where 80\% of the time is used for training ResiDualGAN and 20\% left is for the segmentation model and output space adaptation. The total time consumption is approximate with DualGAN\cite{yi_dualgan_2017} and CycleGAN\cite{zhu_unpaired_2017} and is more efficient than MUCSS\cite{li_learning_2021} (about 30 hours), which needs to generate pseudo labels and perform self-training. 
\subsubsection{Stage A}
We set $k=1$, $\lambda_{adv}=1$, and $\lambda_{cycle}=10$ for training of the ResiDualGAN. Adam\cite{diederik_adam_2017} with $\beta=\left(0.5,0.999\right)$ is adopted as the optimizer for ResiGenerators, while RMSProp\cite{tieleman_lecture_2012} with $\alpha=0.99$ for discriminators. The learning rates for all ResiGenerators and discriminators are set as 0.0005. The batch size is set as 1, where we randomly select images from the source domain and the target domain for training. For every 5 iterations of training for discriminators, 1 iteration of training for ResiGenerators is performed. Finally, a total of 100 epochs are trained.
\subsubsection{Stage B}
We set $\lambda_{seg}=1$ and $L_{adv_o}=0.02$ for the training of Stage B. Adam with $\beta=(0.9,0.999)$ is adopted as the optimizer for semantic segmentation model $f_T$ and output space discriminator $D_{out}$. The initial learning rates are both set as 0.0002. We dynamically adjust the learning rate of  $f_T$ by multiplying 0.5 when the metrics has stopped ascending. The batch size is set as 16.

\section{Experimental Results}

\subsection{Datasets}
\begin{figure}[ht]
    \centering
    \includegraphics[width=.48\textwidth]{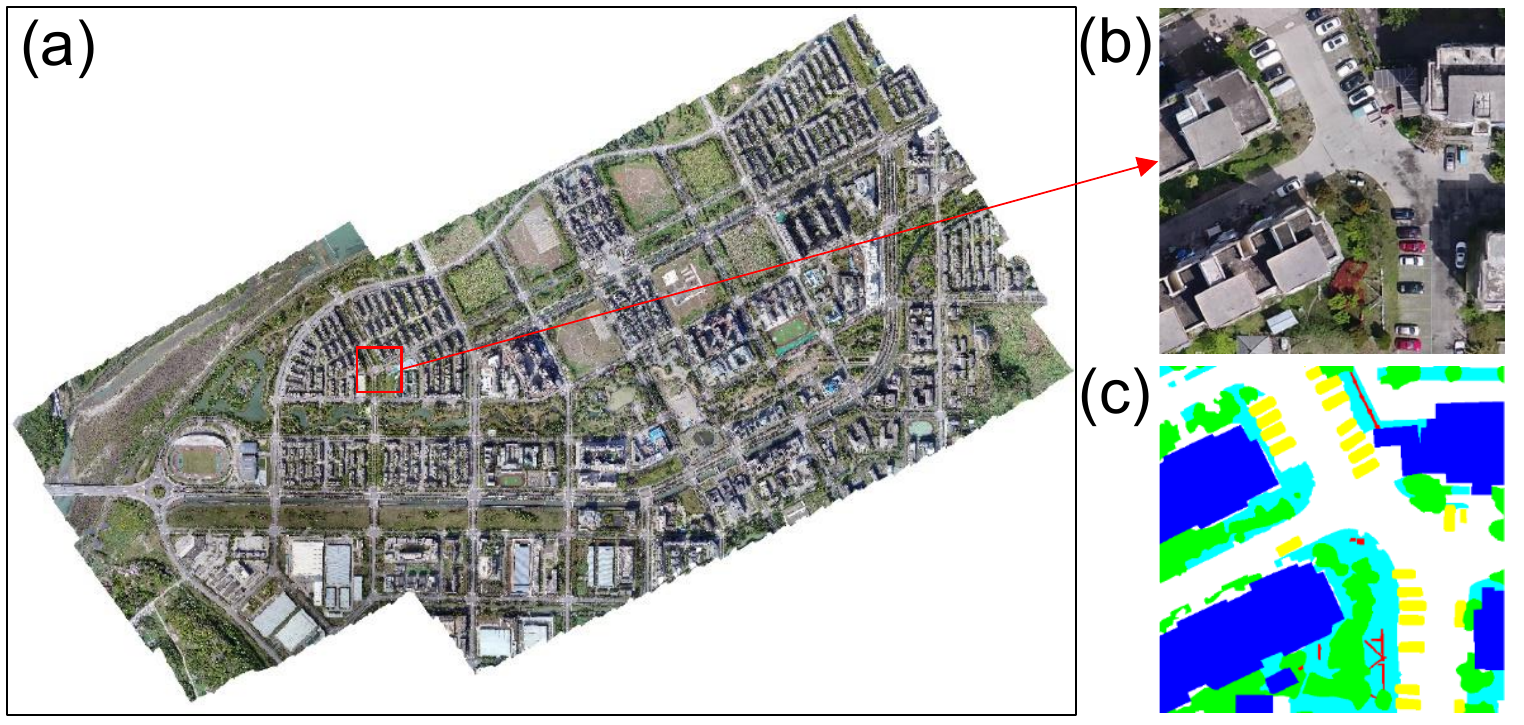}
    \caption{The overview of BC403 dataset (a). True orthophotos with size of 768 × 768 (b) and its semantic labels (c).}
    \label{fig:BC403}
\end{figure}
\begin{table*}[]
\centering
\renewcommand\arraystretch{1.3}
\caption{\centering The quantitative results of the cross-domain semantic segmentation from PotsdamIRRG to Vaihingen. Results marked with * are from the original paper.}
\Huge
\label{tab:PotsdamIRRG2Vaihingen}
\resizebox{\textwidth}{!}{%
\begin{tabular}{@{}ccccccccccccccc@{}}
\toprule[5pt]
Methods &
  \multicolumn{2}{c}{\begin{tabular}[c]{@{}c@{}}Background/\\ Clutter\end{tabular}} &
  \multicolumn{2}{c}{\begin{tabular}[c]{@{}c@{}}Impervious \\ Surface\end{tabular}} &
  \multicolumn{2}{c}{Car} &
  \multicolumn{2}{c}{Tree} &
  \multicolumn{2}{c}{\begin{tabular}[c]{@{}c@{}}Low \\ Vegetation\end{tabular}} &
  \multicolumn{2}{c}{Building} &
  \multicolumn{2}{c}{Overall} \\ \toprule[5pt]
                 & IoU   & F1\-score & IoU   & F1\-score & IoU   & F1\-score & IoU   & F1\-score & IoU   & F1\-score & IoU   & F1\-score & IoU   & F1\-score \\ \midrule[4pt]
Baseline (DeeplabV3\cite{chen2017rethinking})         & 2.12  & 4.01      & 47.68 & 64.47     & 20.39 & 33.62     & 51.37 & 67.81     & 30.25 & 46.38     & 65.74 & 79.28     & 36.26 & 49.26     \\ \midrule[4pt]
CycleGAN\cite{zhu_unpaired_2017}         & 6.93  & 9.95      & 57.41 & 72.67     & 20.74 & 33.46     & 44.31 & 61.08     & 35.60 & 52.17     & 65.71 & 79.12     & 38.45 & 51.41     \\
DualGAN\cite{yi_dualgan_2017}          & 7.70  & 11.12     & 57.98 & 73.04     & 25.20 & 39.43     & 46.12 & 62.79     & 33.77 & 50.00     & 64.24 & 78.02     & 39.17 & 52.40     \\
AdaptSegNet\cite{tsai_learning_2018}      & 5.84  & 9.01      & 62.81 & 76.88     & 29.43 & 44.83     & 55.84 & 71.45     & 40.16 & 56.87     & 70.64 & 82.66     & 44.12 & 56.95     \\
MUCSS\cite{li_learning_2021}           & 10.82 & 14.35     & 65.81 & 79.03     & 26.19 & 40.67     & 50.60 & 66.88     & 39.73 & 56.39     & 69.16 & 81.58     & 43.72 & 56.48     \\
FSDAN\cite{ji_generative_2021}*            & 10.00 & /         & 57.4 & /         & 37.0 & /         & 58.40 & /         & 41.7 & /         & 57.8 & /         & 43.70 & /         \\
LuBin's\cite{lubin}*         & \textbf{19.60}  & \textbf{32.80}      & 65.00 & 78.80     & 39.60 & 56.70     & 54.80 & 70.80     & 36.20 & 53.20     & 76.00 & 86.40     & 48.50 & 63.10     \\ \midrule[4pt]
ResiDualGAN      & 8.20  & 13.71     & 68.15 & 81.03     & 49.50 & 66.06     & 61.37 & 76.03     & 40.82 & 57.86     & 75.50 & 86.02     & 50.59 & 63.45     \\
ResiDualGAN +OSA & 11.64 & 18.42     & \textbf{72.29}  & \textbf{83.89}      & \textbf{57.01}  & \textbf{72.51}      & \textbf{63.81}  & \textbf{77.88}      & \textbf{49.69}  & \textbf{66.29}      & \textbf{80.57}  & \textbf{89.23}      & \textbf{55.83}  & \textbf{68.04}     \\ \bottomrule[5pt]
\end{tabular}
}
\end{table*}
\begin{table*}[]
\centering
\renewcommand\arraystretch{1.3}
\caption{\centering The quantitative results of the cross-domain semantic segmentation from PotsdamRGB to Vaihingen. Results marked with * are from the original paper.}
\Huge
\label{tab:PotsdamRGB2Vaihingen}
\resizebox{\textwidth}{!}{%
\begin{tabular}{@{}ccccccccccccccc@{}}
\toprule[5pt]
Methods &
  \multicolumn{2}{c}{\begin{tabular}[c]{@{}c@{}}Background/\\ Clutter\end{tabular}} &
  \multicolumn{2}{c}{\begin{tabular}[c]{@{}c@{}}Impervious \\ Surface\end{tabular}} &
  \multicolumn{2}{c}{Car} &
  \multicolumn{2}{c}{Tree} &
  \multicolumn{2}{c}{\begin{tabular}[c]{@{}c@{}}Low \\ Vegetation\end{tabular}} &
  \multicolumn{2}{c}{Building} &
  \multicolumn{2}{c}{Overall} \\ \toprule[5pt]
                 & IoU   & F1\-score & IoU   & F1\-score & IoU   & F1\-score & IoU   & F1\-score & IoU   & F1\-score & IoU   & F1\-score & IoU   & F1\-score \\ \midrule[4pt]
Baseline ((DeeplabV3\cite{chen2017rethinking})         & 1.81  & 3.43      & 46.29 & 63.17     & 13.53 & 23.70     & 40.23 & 57.27     & 14.57 & 25.39     & 60.78 & 75.56     & 29.53 & 41.42     \\ \midrule[4pt]
CycleGAN\cite{zhu_unpaired_2017}         & 2.03  & 3.14      & 48.48 & 64.99     & 25.99 & 40.57     & 41.97 & 58.87     & 23.33 & 37.50     & 64.53 & 78.26     & 34.39 & 47.22     \\
DualGAN\cite{yi_dualgan_2017}          & 3.97  & 6.67      & 49.94 & 66.23     & 20.61 & 33.18     & 42.08 & 58.87     & 27.98 & 43.40     & 62.03 & 76.35     & 34.44 & 47.45     \\
AdaptSegNet\cite{tsai_learning_2018}      & 6.49  & 9.82      & 55.70 & 71.24     & 33.85 & 50.05     & 47.72 & 64.31     & 22.86 & 36.75     & 65.70 & 79.15     & 38.72 & 51.89     \\
MUCSS\cite{li_learning_2021}            & 8.78  & 12.78     & 57.85 & 73.04     & 16.11 & 26.65     & 38.20 & 54.87     & 34.43 & 50.89     & 71.91 & 83.56     & 37.88 & 50.30     \\
LuBin's\cite{lubin} *         & \textbf{10.80} & \textbf{19.40}     & \textbf{62.40} & \textbf{76.90}     & 38.90 & 56.00     & 53.90 & 70.00     & \textbf{35.10} & \textbf{51.90}     & 74.80 & 85.60     & 46.00 & \textbf{60.00}     \\ \midrule[4pt]
ResiDualGAN      & 8.80  & 13.90     & 52.01 & 68.35     & 42.58 & 59.58     & 59.88 & 74.87     & 31.42 & 47.69     & 69.61 & 82.04     & 44.05 & 57.74     \\
ResiDualGAN +OSA & 9.76  & 16.08     & 55.54 & 71.36     & \textbf{48.49} & \textbf{65.19}     & \textbf{57.79} & \textbf{73.21}     & 29.15 & 44.97     & \textbf{78.97} & \textbf{88.23}     & \textbf{46.62} & 59.84    \\  
\bottomrule[5pt]
\end{tabular}
}
\end{table*}
To fully verify the effectiveness of the proposed UDA method, three VHR RS datasets are introduced in the experiment, namely Potsdam dataset, Vaihingen dataset, and BC403 dataset. 

The first two datasets belong to the ISPRS 2D open-source RS semantic segmentation benchmark dataset\cite{potsdam_vaihingen}, all images in both datasets are produced into true orthophotos (TOPs), with annotations for 6 ground classes: clutter/background, impervious surfaces, car, tree, low vegetation, and building. The Potsdam dataset contains three different band modes: IR-R-G(three channels), R-G-B(three channels), and IR-R-G-B(four channels). IR-R-G and R-G-B are exploited in the following experiments that are abbreviated as PotsdamIRRG and PotsdamRGB, which both consist of 38 VHR TOPs with a fixed size of 6000×6000 pixels and 5cm spatial resolution. The Vaihingen dataset contains only one band mode: IR-R-G(three channels) and consists of 33 TOPs in which every TOP contains 2000 × 2000 pixels, with a resolution of 9cm. To conform to the constraints of Equation \eqref{formu:1}, we clip images of Potsdam into the size of 896 × 896 and images of Vaihingen into the size of 512 × 512, where the numbers 896 and 512 are specifically set for facilitating downsampling operation in the CNN. Eventually, 1296 images for PotsdamIRRG and PotsdanRGB and 1696 images for Vaihingen, in which 440 images of Vaihingen are validation datasets, are obtained.

\begin{table*}[]
\centering
\renewcommand\arraystretch{1.3}
\caption{\centering The quantitative results of the cross-domain semantic segmentation from PotsdamIRRG to BC403.}
\Huge
\label{tab:PotsdamIRRG2BC403}
\resizebox{\textwidth}{!}{%
\begin{tabular}{@{}ccccccccccccccc@{}}
\toprule[5pt]
Methods &
  \multicolumn{2}{c}{\begin{tabular}[c]{@{}c@{}}Background/\\ Clutter\end{tabular}} &
  \multicolumn{2}{c}{\begin{tabular}[c]{@{}c@{}}Impervious \\ Surface\end{tabular}} &
  \multicolumn{2}{c}{Car} &
  \multicolumn{2}{c}{Tree} &
  \multicolumn{2}{c}{\begin{tabular}[c]{@{}c@{}}Low \\ Vegetation\end{tabular}} &
  \multicolumn{2}{c}{Building} &
  \multicolumn{2}{c}{Overall} \\ \toprule[5pt]
                 & IoU   & F1\-score & IoU   & F1\-score & IoU   & F1\-score & IoU   & F1\-score & IoU   & F1\-score & IoU   & F1\-score & IoU   & F1\-score \\ \midrule[4pt]
Baseline (DeeplabV3\cite{chen2017rethinking})         & 2.74  & 5.26      & 27.35 & 42.44     & 35.68 & 52.22     & 43.18 & 60.02     & 11.43 & 19.99     & 60.51 & 75.20     & 30.15 & 42.52     \\ \midrule[4pt]
CycleGAN \cite{zhu_unpaired_2017}        & 4.79  & 6.54      & 68.37 & 79.65     & 47.99 & 54.76     & 22.69 & 45.76     & 22.68 & 42.05     & 72.57 & 80.85     & 39.85 & 51.60     \\
DualGAN\cite{yi_dualgan_2017}          & 3.52  & 8.78      & 66.48 & 80.97     & 39.26 & 64.35     & 30.58 & 36.68     & 28.03 & 35.73     & 68.53 & 83.63     & 39.40 & 51.69     \\
AdaptSegNet\cite{tsai_learning_2018}      & 4.57  & 8.17      & 56.37 & 72.03     & 50.90 & 66.95     & \textbf{52.01} & \textbf{67.87}     & 14.70 & 25.06     & 74.38 & 85.17     & 42.15 & 54.21     \\
MUCSS\cite{li_learning_2021}            & 3.13  & 5.86      & 71.83 & 83.35     & 27.72 & 41.62     & 32.53 & 48.17     & 27.95 & 42.08     & 78.82 & 87.82     & 40.33 & 51.48     \\ \midrule[4pt]
ResiDualGAN      & \textbf{13.79} & \textbf{23.72}     & 72.26 & 83.64     & 61.06 & 75.69     & 46.56 & 62.76     & 33.73 & 49.67     & 76.08 & 86.15     & 50.58 & 63.61     \\
ResiDualGAN +OSA & 13.25 & 23.03     & \textbf{75.51} & \textbf{85.95}     & \textbf{61.32} & \textbf{75.87}     & 51.22 & 67.18     & \textbf{35.35} & \textbf{51.33}     & \textbf{82.51} & \textbf{90.24}     & \textbf{53.19} & \textbf{65.60}      \\
\bottomrule[5pt]
\end{tabular}
}
\end{table*}
\begin{figure*}[ht]
    \centering
    \setlength{\abovecaptionskip}{0pt}
    \setlength{\belowcaptionskip}{0pt}
    \includegraphics[width=0.9\textwidth, height=0.33\textheight]{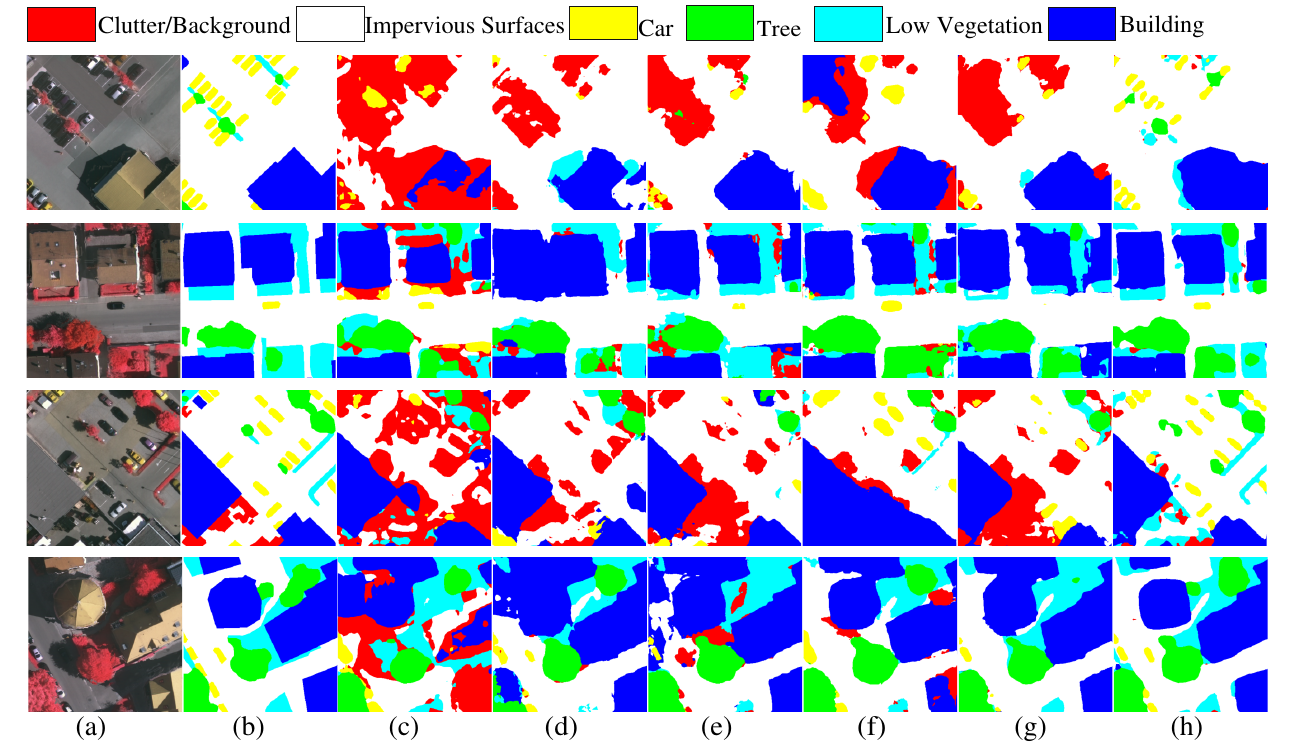}
    \caption{The qualitative results of the cross-domain semantic segmentation from PotsdamIRRG to Vaihingen. (a) Target images. (b) Labels. (c) Baseline (DeeplabV3\cite{chen2017rethinking}). (d) CycleGAN\cite{zhu_unpaired_2017}. (e) DualGAN\cite{yi_dualgan_2017}. (f) AdaptSegNet\cite{tsai_learning_2018}. (g) MUCSS\cite{li_learning_2021}. (h) ResiDualGAN+OSA(ours). }
    \label{fig:PotsdamIRRG2Vaihingen}
\end{figure*}

The third dataset (BC403) is manually annotated on our own. This dataset is constructed using VHR drone imagery obtained from Beichuan County area with a spatial resolution of 7 cm. Beichuan County is located in southwestern China, under the jurisdiction of Mianyang City, Sichuan Province, and this dataset is collected in April 2018 and is part of the Wenchuan Earthquake 10th Anniversary UAV dataset. Beichuan County has a subtropical monsoonal humid climate and has architectural features typical of Chinese counties, which is certain different from the above two datasets in terms of image feature characteristics. The original image size of the dataset is 27953 × 43147 Fig. \ref{fig:BC403}(a). The down-sampled RS images are seamlessly cropped into 1584 tiles with 768 × 768 pixels with 30\% overlapping Fig. \ref{fig:BC403}(b). Images in this dataset are provided with their semantic labels Fig. \ref{fig:BC403}(c), including six classes of ground objects: clutter/background, impervious surfaces, car, tree, low vegetation, and building, as in the Potsdam and Vaihingen dataset. We select 20\% images as the validation dataset and others for the test dataset. And the test dataset doesn't overlap with the validation dataset.

\subsection{Experimental Settings}
We design three cross-domain tasks to simulate situations that might
be encountered in practical applications using the above datasets:
\begin{enumerate}
    \item IR-R-G to IR-R-G: PotsdamIRRG to Vaihingen. A commonly used benchmark for evaluating models. 
    \item R-G-B to IR-R-G: PotsdamRGB to Vaihingen. Another commonly used benchmark for evaluating models. 
    \item IR-R-G to RGB: PotsdamIRRG to BC403. Instead of using PotsdamRGB as the target dataset where channels of R and G are identical with PotsdamIRRG, we use our annotated BC403 dataset to perform this cross-domain task. 
\end{enumerate}
\begin{figure*}[ht]
    \centering
    \setlength{\abovecaptionskip}{0pt}
    \setlength{\belowcaptionskip}{0pt}
    \includegraphics[width=0.9\textwidth, height=0.35\textheight]{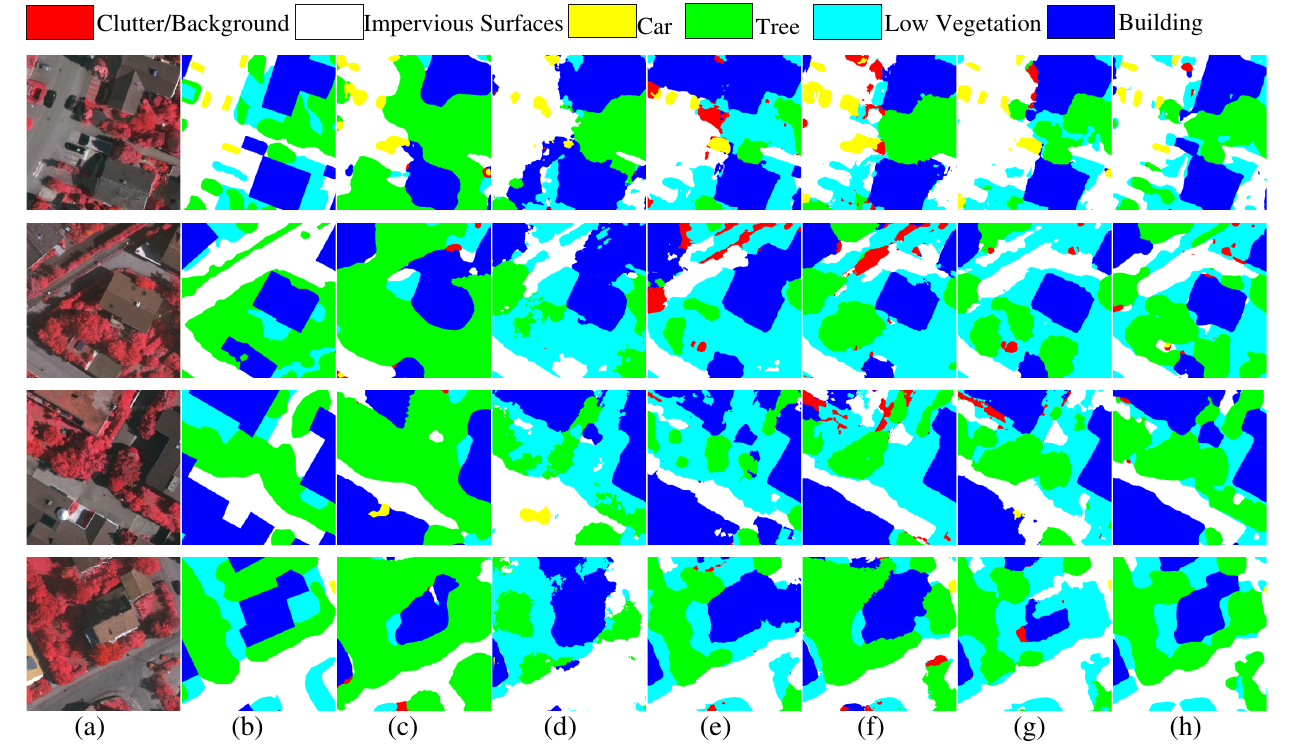}
    \caption{The qualitative results of the cross-domain semantic segmentation from PotsdamRGB to Vaihingen. (a) Target images. (b) Labels. (c) Baseline (DeeplabV3\cite{chen2017rethinking}). (d) CycleGAN\cite{zhu_unpaired_2017}. (e) DualGAN\cite{yi_dualgan_2017}. (f) AdaptSegNet\cite{tsai_learning_2018}. (g) MUCSS\cite{li_learning_2021}. (h) ResiDualGAN+OSA(ours). }
    \label{fig:PotsdamRGB2Vaihingen}
\end{figure*}
\begin{figure*}[ht]
    \centering
    \setlength{\abovecaptionskip}{0pt}
    \setlength{\belowcaptionskip}{0pt}
    \includegraphics[width=0.9\textwidth, height=0.35\textheight]{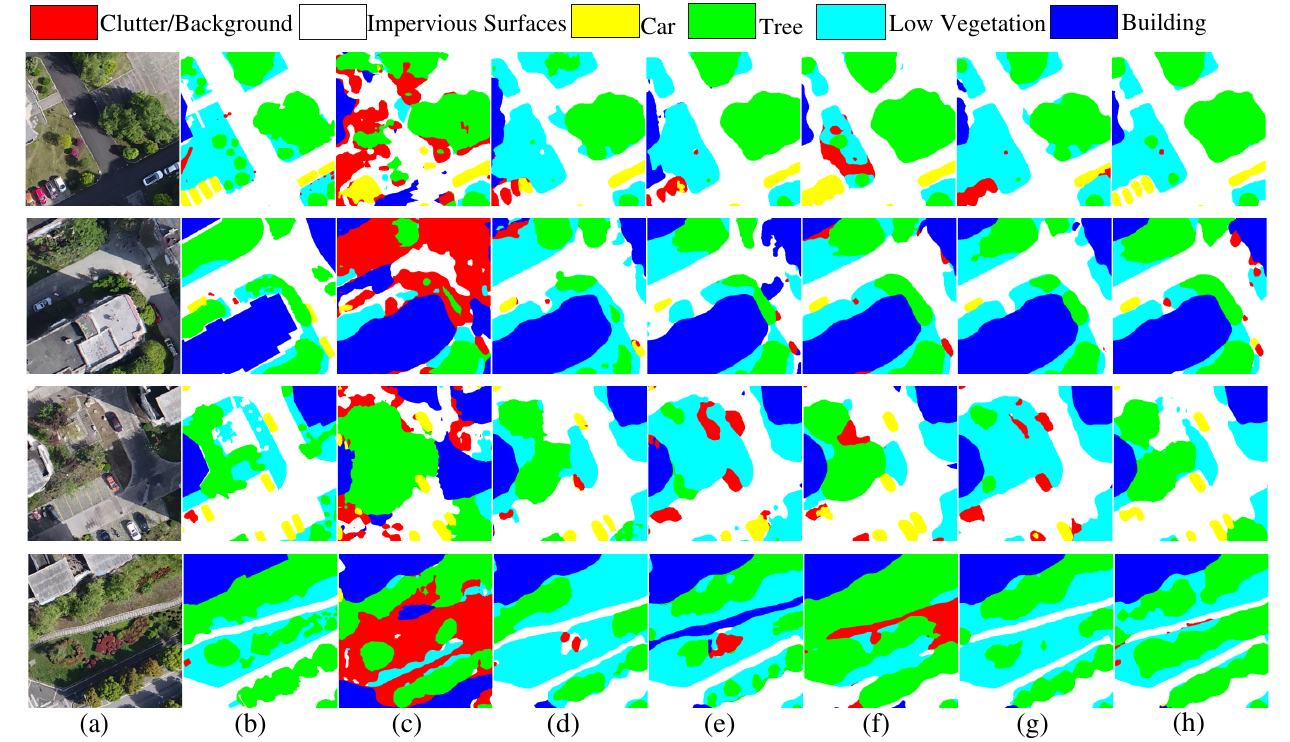}
    \caption{The qualitative results of the cross-domain semantic segmentation from PotsdamIRRG to BC403. (a) Target images. (b) Labels. (c) Baseline (DeeplabV3\cite{chen2017rethinking}). (d) CycleGAN\cite{zhu_unpaired_2017}. (e) DualGAN\cite{yi_dualgan_2017}. (f) AdaptSegNet\cite{tsai_learning_2018}. (g) MUCSS\cite{li_learning_2021}. (h) ResiDualGAN+OSA(ours). }
    \label{fig:PotsdamIRRG2BC403}
\end{figure*}

\subsection{Evaluation Metrics}
To facilitate comparing with different methods, IoU and F1-score are employed as metrics in this paper. For every class in 6 different ground classes, the formulation of IoU can be written as:
\begin{equation}
    IoU=\frac{\left|A\cap B\right|}{\left| A\cup B \right|}
    \label{formu:11}
\end{equation}
Where A is the ground truth while B is predictions. After calculations of IoU for 6 classes, mIoU can be obtained which is the mean of IoU for every class. And the F1-score can be written as:
\begin{equation}
F1\text{-}score=\frac{2\times Precision\times Recall}{Precision+Recall}
\label{formu:12}
\end{equation}

\subsection{Compared with the State-of-the-art Methods}

For a comprehensive comparative analysis, we keep all of the settings of networks consistent with ours and hyperparameters in other methods as optimal as possible. Four state-of-the-art methods are used for comparison, i.e., CycleGAN\cite{zhu_unpaired_2017}, DualGAN\cite{yi_dualgan_2017}, AdaptSegNet\cite{tsai_learning_2018}, and MUCSS\cite{li_learning_2021}, where the former two are image-to-image translation methods, AdaptSegNet is an adversarial discriminative method and MUCSS combines DualGAN with self-training strategies together for cross-domain semantic segmentation tasks. In addition, we added the results of LuBin's research\cite{lubin} in the task of cross-domain semantic segmentation from PotsdamIRRG to Vaihingen and PotsdamRGB to Vaihingen; meanwhile, the method of FSDAN\cite{ji_generative_2021} was compared in the processing of PotsdamIRRG to Vaihingen.

Both the quantitative and qualitative results show the superiority of the proposed methods. Table \ref{tab:PotsdamIRRG2Vaihingen} and Fig. \ref{fig:PotsdamIRRG2Vaihingen} show the quantitative and qualitative segmentation results of PotsdamIRRG to Vaihingen respectively, Table \ref{tab:PotsdamRGB2Vaihingen} and Fig. \ref{fig:PotsdamRGB2Vaihingen} are PotsdamRGB to Vaihingen while Table \ref{fig:PotsdamIRRG2BC403} and Fig. \ref{fig:PotsdamIRRG2BC403} are PotsdamIRRG to BC403. After the OSA, we finally get the mIoU and F1-score of segmentation results of 55.83\% and 68.04\% from PotsdamIRRG to Vaihingen, increasing by 7.33\% and 4.94\% respectively compared with other methods. For PotsdamIRRG to BC403, we get results of 53.19\% and 65.60\%, increasing by 11.04\% and 11.39\% . For the result of PotsdamRGB to Vaihingen, the improvement of mIOU is not obvious, only 0.62\%, and the value of the F1-score (59.84\%) is slightly lower than the results of other studies (60.00\%). However, it can be observed that the improvement in car class is significant (mIoU from 38.9\% of MUCSS to 48.49\% of ours), while the improvement in low vegetation is deficient (mIoU from 35.10\% of MUCSS to 29.15\% of ours).  In section \ref{resizer_module}, we will discuss the principal reason for these imbalanced improvements.

\section{Discussion}

\subsection{Hyperparameters Settings}

\begin{table}[]
\centering
\renewcommand\arraystretch{1.5}
\caption{\centering Evaluation results of ResiDualGAN under different hyperparameters settings. The results are get from the task of cross-domain semantic segmentation from PotsdamIRRG to Vaihingen and evaluate on the \textbf{validation part} of Vaihingen. The mIoU and F1 are overall IoU and overall F1-score respectively. }
\label{tab:hyper}
\resizebox{.48\textwidth}{!}{%
\begin{tabular}{@{}lccc@{}}
\toprule[1.5pt]
\multicolumn{2}{c}{Hyperparameters Settings} & mIoU & F1 \\ 
\toprule[1.5pt]
\multirow{3}{*}{ $L_{adv},L_{cyc}=(1,10)$} & $k=0.5$ & 53.29 & 66.09 \\
 & \underline{$k=1$} & \textbf{56.80} & \textbf{68.46} \\
 & $k=2$ & 55.10 & 67.39 \\
 \midrule[1pt]
\multirow{10}{*}{$k=1$} & $L_{adv}, L_{cyc}=(1,1)$ & 56.00 & 68.24 \\
 &  $L_{adv}, L_{cyc}=(1,5)$ & 56.71 & 69.10 \\
 &\underline{$L_{adv}, L_{cyc}=(1,10)$} & \textbf{56.80} & \textbf{68.46} \\
 &  $L_{adv}, L_{cyc}=(1,20)$ & 54.79 & 67.34 \\
 &  $L_{adv}, L_{cyc}=(5,1)$ & 54.43 & 67.33 \\
 &  $L_{adv}, L_{cyc}=(5,5)$ & 52.69 & 65.02 \\
 &  $L_{adv}, L_{cyc}=(5,10)$ & 55.73 & 68.26 \\
 &  $L_{adv}, L_{cyc}=(10,1)$ & 53.39 & 65.51 \\
 &  $L_{adv}, L_{cyc}=(10,5)$ & 55.12 & 67.10 \\
 &  $L_{adv}, L_{cyc}=(10,10)$ & 54.98 & 67.56 \\ 
 \bottomrule[1.5pt]
\end{tabular}%
}
\end{table}
To boost the performance of our model, we evaluate the proposed method (ResiDualGAN+OSA) on the evaluation datasets of Vaihingen under the task of transferring the segmentation model from PotsdamIRRG to Vaihingen. The grid search method is used to find the optimal hyperparameters combination. Table  \ref{tab:hyper} shows the results of the grid search. Firstly, for a given $L_{adv}, L_{cyc}=(1,10)$, we evaluate model’s performance under different $k$ settings. When $k=1$, the best result is reached with mIoU=56.80\% and overall F1-score=68.46\%. Secondly, fixed $k=1$, we evaluate how the settings of $L_{adv}, L_{cyc}$ effect model’s performance. We split the grid of the search space into $\{1,5,10\}\times \{1,5,10\}$, where $\times$ is the Cartesian product. The best result is got when $L_{adv}, L_{cyc}=(1,10)$. Eventually, we set the hyperparameters of ResiDualGAN as $k, L_{adv}, L_{cyc}=(1,1,10)$. 

In addition, we can observe that the results don't fluctuate too much under different hyperparameters settings (max of mIou - min of mIou = 56.80\% - 52.69\% = 4.11\%), which demonstrates the stability of our model under different hyperparameters settings. 

\subsection{Image Translation}
\begin{figure}[ht]
    \centering
    \setlength{\abovecaptionskip}{0pt}
    \setlength{\belowcaptionskip}{0pt}
    \includegraphics[width=0.48\textwidth]{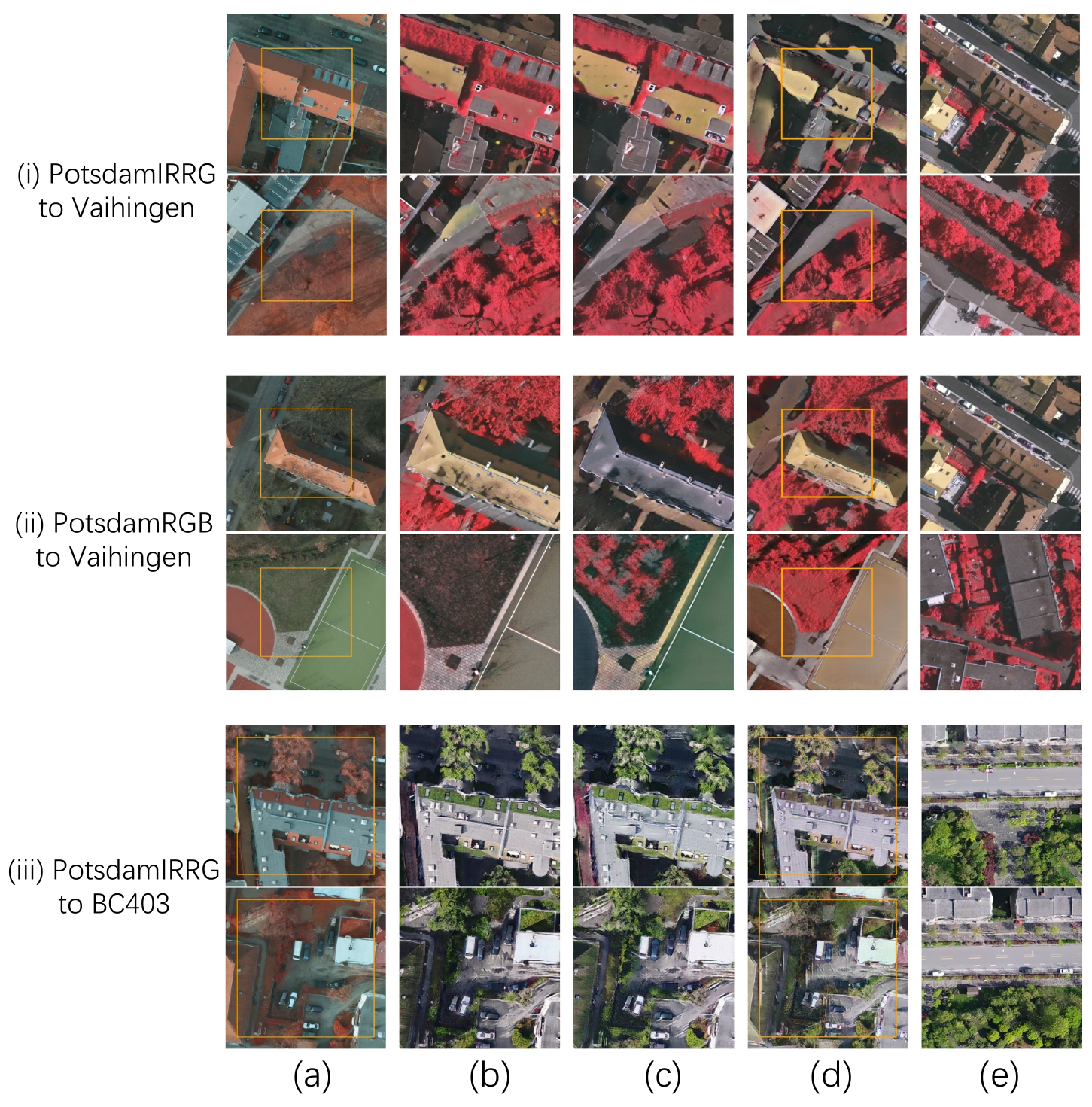}
    \caption{Results of image translation from (i) PotsdamIRRG to Vaihingen, (ii) PotsdamRGB to Vaihingen and (iii) PotsdamIRRG to BC403. (a) Input images. (b) CycleGAN\cite{zhu_unpaired_2017}. (c) DualGAN\cite{yi_dualgan_2017}. (d) ResiDualGAN. (e) Target images. The area with orange rectangle in (a) and (d) is used for image translation in (b) and (c), where in (i) and (ii) the size of (a) is $896\times896$ and the orange rectangle of (a) is $512\times512$ as well as the size of (b), (c), (d), and (e) to conform the Equation\ref{formu:1}. In (iii), the size of (a) is $896\times896$ and the orange rectangle of (a) is $768\times768$ as well as the size of (b), (c), (d), and (e). }
    \label{fig:transfer}
\end{figure}
\begin{figure*}[ht]
    \centering
    \setlength{\abovecaptionskip}{0pt}
    \setlength{\belowcaptionskip}{0pt}
    \includegraphics[width=1.0\textwidth]{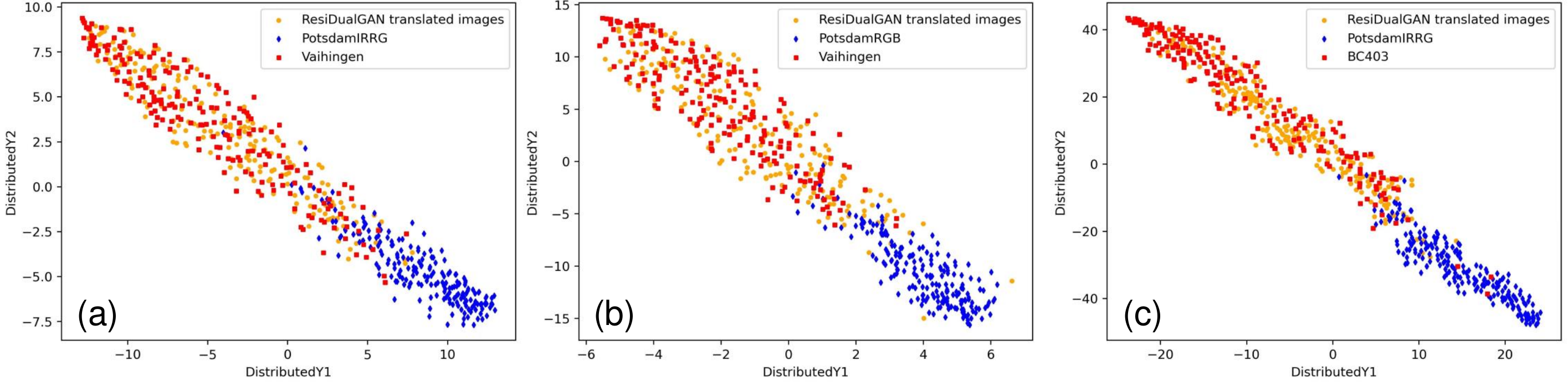}
    \caption{Visualization of the t-sne\cite{van2008visualizing} results of ResiDualGAN. Every point in the figure refers to the t-sne dimension reduction result of the feature of an image.  The feature is obtained from the encoder of a ResNet-18\cite{he_deep_2016} network. The orange dots refer to features of images generated by ResiDualGAN under the image translation tasks from (a) PotsdamIRRG to Vaihingen, (b) PotsdamRGB to Vaihingen, (c) PotsdamIRRG to BC403. The other points refer to features of images from PotsdamIRRG/PotsdamRGB/Vaihingen/BC403 .}
    \label{fig:tsne}
\end{figure*}
Fig. \ref{fig:transfer} shows the image translation results of ResiDualGAN. The translation of the tan roof in Fig. \ref{fig:transfer}(i-a) is a tough problem for the existing GANs, where the tan roof is likely to be translated as the low vegetation e.g., Fig. \ref{fig:transfer}(i-b) and Fig. \ref{fig:transfer}(i-c). The ResiDualGAN avoids this problem as shown in Fig. \ref{fig:transfer}(i-d), where the tan roof is translated into yellow which is corresponding to Vaihingen in Fig. \ref{fig:transfer}(i-e). Though great improvement has been made, however, the translated results of ResiDualGAN are still visually unfriendly, where the color of the imperious surface is translated into yellow, the shadow is too thick to recognize the objects below the shadow, and so on. Fortunately, though visually unfriendly, the translated results of ResiDualGAN are suited for the training in Stage B. The state-of-the-art segmentation performance fully proves the superiority of ResiDualGAN in RS images' cross-domain semantic segmentation tasks. 

Figure. \ref{fig:tsne} shows the t-sne\cite{van2008visualizing} visualized result of image translation. Initially, we trained a classification network to distinguish images from PotsdamIRRG and Vaihingen (PotsdamRGB and Vaihingen, or PotsdamIRRG and BC403). We select ResNet-18\cite{he_deep_2016} as our backbone of classification. Then we feed forward the translated images into networks and get the features extracted by ResNet-18. Finally, the features are visualized using t-sne. The visualization result shows that ResiDualGAN well matches the data distribution of the source domain data with the target domain data. The feature distribution of most of translated images is similar to the target domain data feature distribution.

\subsection{Ablation Study}
\begin{table}[]
\centering
\caption{\centering Ablation study for ResiDualGAN. The results are get from the task of cross-domain semantic segmentation from PotsdamIRRG to Vaihingen and evaluate on the \textbf{test part} of Vaihingen. The mIoU and F1 are overall IoU and overall F1-score respectively. }
\label{tab:ablation}
\renewcommand\arraystretch{1.5}
\resizebox{.48\textwidth}{!}{%
\begin{tabular}{@{}cccc@{}}
\toprule[1.5pt]
Experiment & Method & mIoU & F1 \\ \toprule[1.5pt]
\multirow{3}{*}{Resize} & No   Resize & 44.97 & 58.51 \\
 & Pre-resize & 53.46 & 66.10 \\
 & \underline{In-network   Resize} & \textbf{55.83} & \textbf{68.04} \\ \midrule[1pt]
\multirow{3}{*}{Resizing Function} & Nearest & 53.86 & 66.30 \\
 & \underline{Bilinear} & \textbf{55.83} & \textbf{68.04} \\
 & Resizer   model & 52.97 & 65.88 \\ \midrule[1pt]
\multirow{3}{*}{Backbone} & ResNet\cite{he_deep_2016} & 52.51 & 65.33 \\
 & LinkNet\cite{chaurasia2017linknet} & 52.22 & 64.84 \\
 & \underline{U-Net\cite{ronneberger_u-net_2015}} & \textbf{55.83} & \textbf{68.04} \\ \midrule[1pt]
\multirow{2}{*}{Residual Connection} & No   Residual & 38.67 & 52.37 \\
 & \underline{Residual   (fixed k)} & \textbf{55.83} & \textbf{68.04} \\ \midrule[1pt]
\multirow{2}{*}{k} & Learnable & 54.05 & 67.02 \\ 
 &\underline{Fixed} & \textbf{55.83} & \textbf{68.04} \\ \bottomrule[1.5pt]
\end{tabular}%
}
\end{table}
\begin{figure}
    \centering
    \includegraphics[width=.48\textwidth]{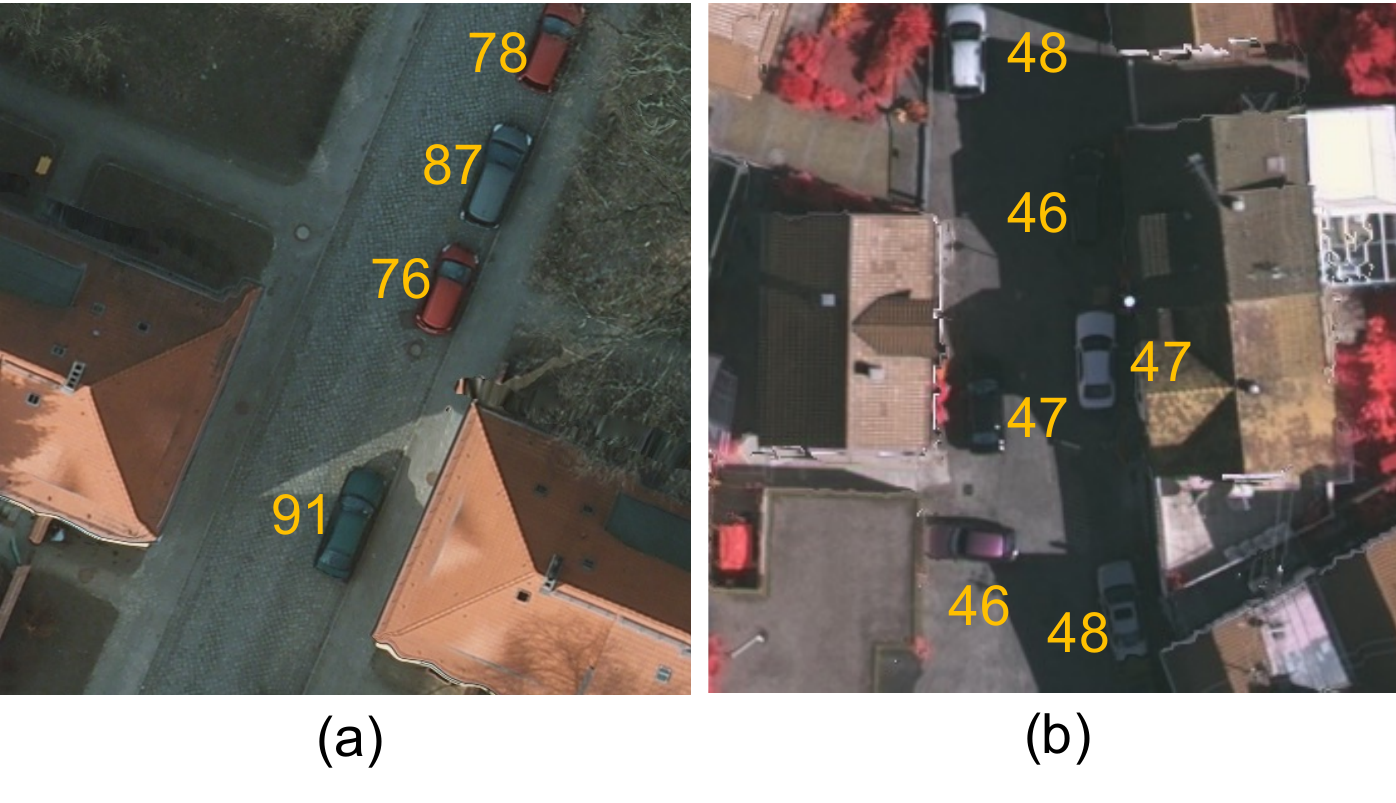}
    \caption{The length of cars in Potsdam (a) and Vaihingen (b) measured by pixel. The number in the figure represents the pixel length of the beside car.}
    \label{fig:cars}
\end{figure}

\subsubsection{Resizer module} \label{resizer_module}
In RS images, some scale-invariant classes (e.g. cars) have a relatively fixed size because of the fixed resolution of RS images. Therefore, if two datasets have different resolutions, the size of cars may be distinct. Fig. \ref{fig:cars} shows that kind of tendency, the size of cars in PotsdamIRRG is close to each other but is always much larger than cars in Vaihingen. 

CNN is a scale-sensitive network\cite{hu2018sinet}. CNN learns to recognize features from the training data and predicts testing data using the knowledge learned from the training data. Consequently, for scale-invariant objects, e.g., cars, scale is a feature that can be learned for CNN. Scale-sensitiveness of CNN brings a great challenge for some CV tasks, such as cars detection from the street scene images\cite{gao2017scale,hu2018sinet}, where cars in that images are in a large variance of scales (as shown in Fig1, cars in the street scene images). However, the variance of scales benefits the UDA tasks. Learning different scale information of a category, CNN possesses the ability to recognize objects with different scales, which benefits the UDA semantic segmentation tasks of car, person, and other scale-invariant classes from GTA5 to Cityscapes. Nevertheless, as mentioned above, the size of scale-invariant classes in an RS dataset are close to each other, which greatly challenges the UDA tasks of RS images.

The resizer module in ResiDualGAN addresses the scale discrepancy problem of two domains. Complying to the Equation \eqref{formu:1}, ResiDualGAN unifies the resolution of target domain images $X_T$ and $ResiG_{S\rightarrow T}(X_S)$ to $r_T$, and resolution of source domain images $X_S$ and $ResiG_{T\rightarrow S}(X_T)$ to $r_S$, addressing the problem that both discriminators $D_S$ and $D_T$ receive two images with different resolution, which may avoid the vanishing gradient problem of discriminators and accelerate the convergence of generators. \par

\begin{figure}[h]
    \centering
    \setlength{\abovecaptionskip}{0pt}
    \setlength{\belowcaptionskip}{0pt}
    \includegraphics[width=.48\textwidth]{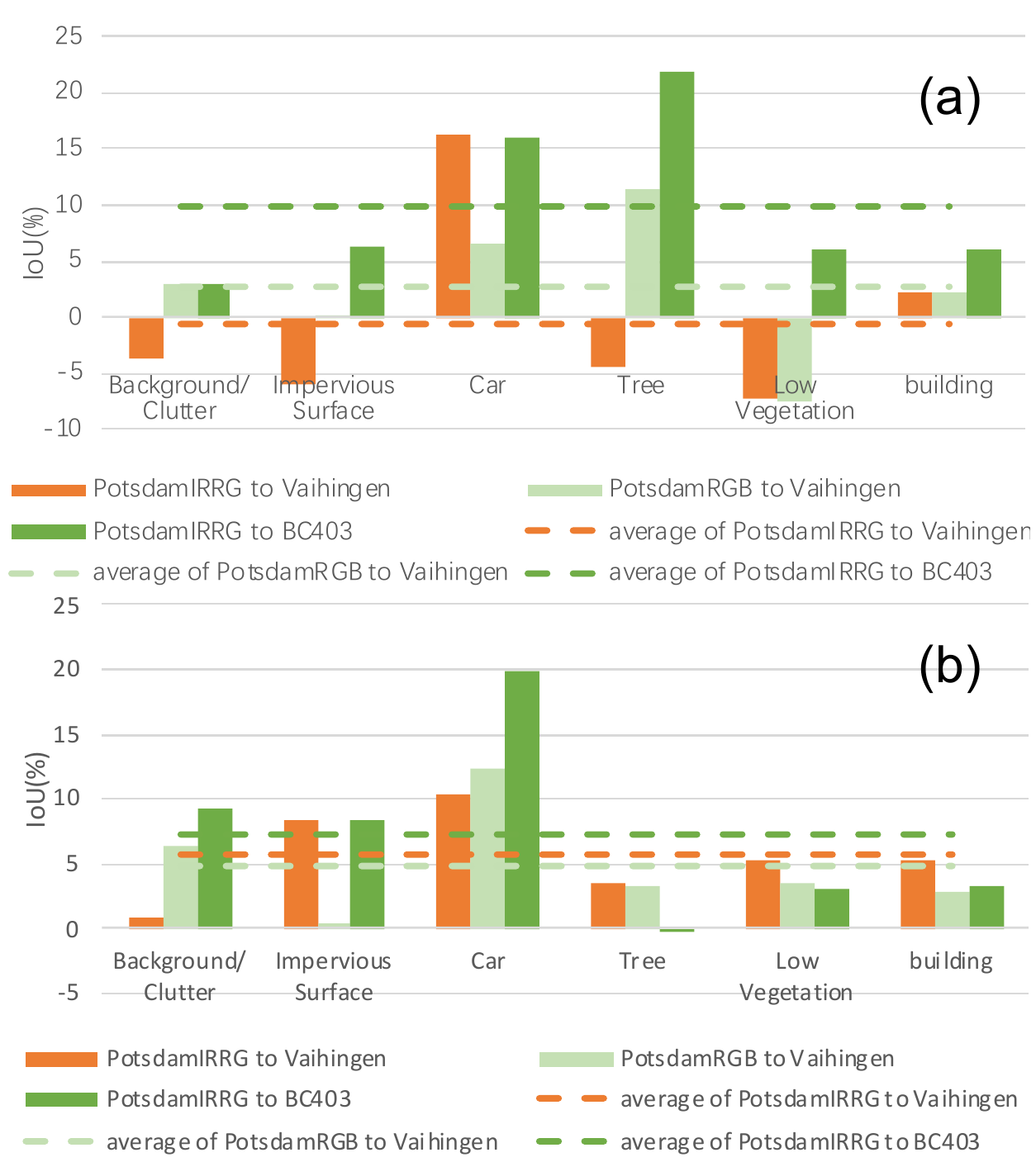}
    \caption{The improvement brought by the resizer module. (a) The difference of mIoU between ResiDualGAN which removes the residual connection and DualGAN. (b) The difference of mIoU between ResiDualGAN and ResiDualGAN which removes the resizer module.}
    \label{fig:sub}
\end{figure}

The resizer module greatly improves the accuracy performance of ResiDualGAN in cross-domain RS image semantic segmentation tasks. If we remove the resizer module, the mIoU drops to 44.97\%, and F1-score drops to 58.51\% (Table \ref{tab:ablation}). Fig. \ref{fig:sub} shows the improvements brought by the resizer module of ResiDualGAN under two pairs of comparison: 1) DualGAN vs ResiDualGAN (No Residual). and 2) ResiDualGAN (No Resizer) vs ResiDuanGAN. ResiDualGAN (No Residual) is just an extension of DualGAN that adds a resizer module after the generator, and ResiDualGAN (No Resizer) only removes the resizer module in ResiGenerator. We perform the experiments on the same hyperparameters settings. The results show that, for scale-invariant classes (e.g. cars), the improvements are much higher than the average. And for scale-invariant classes (e.g. low vegetation), the improvements are under the average. 

Additionally, in-network resizing also affects the performance. Previous works \cite{ji_generative_2021, wittich2021appearance} use resizing function as a pre-processing for input data, which will lead to information loss. An in-network resizer module adapts itself while resizing images, bringing better performance. Table \ref{tab:ablation} shows the experimental results in which a pre-processing resizing operation reduces the mIoU from 55.83\% to 53.46\%, and the F1 from 68.04\% to 66.10\%, which demonstrates the superiority of our method.

\subsubsection{Resizing Function} \label{resizer}
Different image resizer methods may affect the on-task performance of networks\cite{talebi_learning_2021}. Consequently, the implementation of the resizer module will have a significant effect on the semantic segmentation results. In this paper, we compare three types of resizing methods, nearest interpolation, bilinear interpolation, and a resizer model which is proposed by \cite{talebi_learning_2021}. The former two methods are linear methods that contain no parameters to be learned, and the last one is a lightweight network that has shown its superiority compared with linear methods on some CV tasks. Table \ref{tab:ablation} shows the experimental results for the optimal resizer module, where the bilinear interpolation gets the highest mIoU and F1-score. The nearest interpolation gets worse results compared to the bilinear interpolation, resulting from the information loss of images. The resizer model shows the worst results, which should be further optimized to adapt the VHR RS images translation tasks better in the future works. As a result, the bilinear interpolation method is selected as the implementation of the resizer module of ResiDualGAN. 

\subsubsection{Backbone}
\begin{figure}[ht]
    \centering
    \setlength{\abovecaptionskip}{0pt}
    \setlength{\belowcaptionskip}{0pt}
    \includegraphics[width=.48\textwidth]{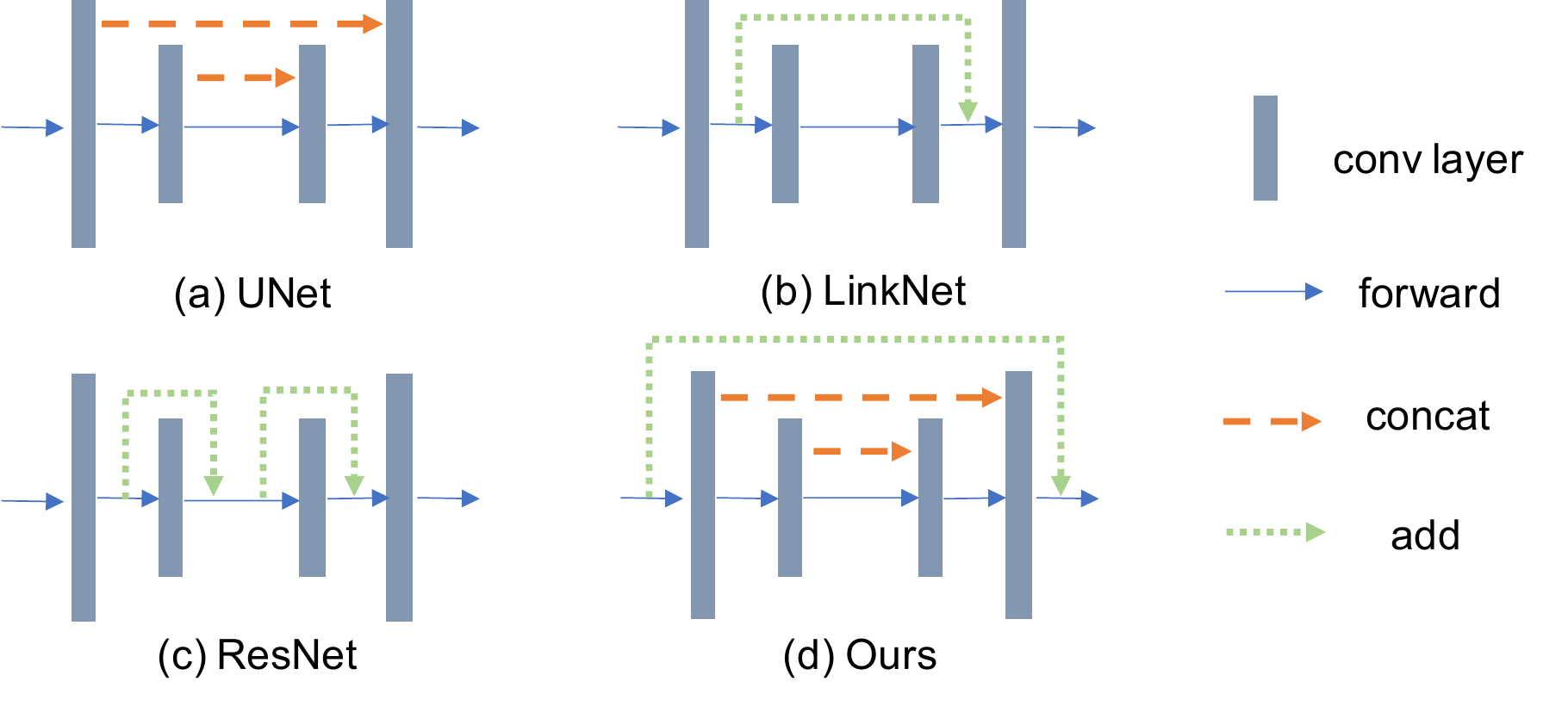}
    \caption{Diagram of the network structure of (a)UNet\cite{ronneberger_u-net_2015}, (b)LinkNet\cite{chaurasia2017linknet}, (c)ResNet\cite{he_deep_2016}, and (d)ResiDualGAN(ours). }
    \label{fig:structure}
\end{figure}
The setting of the backbone of the generator will affect the segmentation model's accuracy. We quantitively compare three CNN-based backbones: U-Net\cite{ronneberger_u-net_2015}, LinkNet\cite{chaurasia2017linknet}, and ResNet\cite{he_deep_2016}. The structure of the three networks is shown in Fig. \ref{fig:structure}. U-Net (Fig. \ref{fig:structure} (a)) is a commonly used backbone in the generation tasks, which connects encoder and decoder layers with feature concatenation. LinkNet (Fig. \ref{fig:structure} (b)) resembles with U-Net which replaces the concat operation as a plus operation between layers. ResNet (Fig. \ref{fig:structure} (c)) utilizes residual connections on a feature level that contributes to building a deeper network. In particular, it is worth noting that the residual connection of ResiDualGAN is totally distinct from it in ResNet. As Fig. \ref{fig:structure} (d) shows, ResiDualGAN merely add the input with the output of the backbone. In ResNet, the skip connection is used to add the input feature with the output feature, where the feature is firstly passed through the encoder and added to the feature with the same channels. The procedure of encoding an image to a feature map brings unnecessary information loss. The optimal way is to add the input image to the output image of the backbone, where the function of the backbone becomes generating a residual item but encoding an image and then decoding to get a new image. The experimental results are shown in Table \ref{tab:ablation}. For fairly comparing, We control the parameters of the three backbones generally equivalent. The quantitive results illustrate that U-Net is the better choice as a backbone for our generative model. The experimental results also show that our residual connection design is much better.

\subsubsection{Residual Connection}
Combining with the resizer module, residual connection plays a pivotal role in achieving state-of-the-art accuracy performance of ResiDualGAN. If we remove the residual connection in our model,  the mIoU drops from 55.83\% to 38.67\%, and the overall F1-score drops from 68.04\% to 52.37\% as Table \ref{tab:ablation} shows. The residual connection retains the original data and avoids the modification of the structure information. Translation between RS datasets is real-to-real where all the pixels are geographically significant. Image-to-image translation GANs like DualGAN are widely utilized in UDA, which not only perform the real-to-real translation but also synthetic-to-real translation, e.g., GTA5 to Cityscapes. However, real-to-real translation is distinct from synthetic-to-real translation. Intuitively, at the procedure of image-to-image translation, the networks should modify the real images less than the synthetic images where the margin distribution between real and real is more closed than synthetic and real\cite{zheng2018t2net}. Meanwhile, it is not expected to modify the structure information of real images which may affect the segmentation performance. Nevertheless, the U-shape network generator of DualGAN is likely to modify the structure information. The residual connection of ResiDualGAN retains the original structure information as much as possible and focuses on the translation of other information, e.g., color, shadow, and so on. As a result, residual connection improves the segmentation performance and is more suitable for RS images translation. 

\subsubsection{Fixed $k$}
$k$ is a vital parameter for ResiDualGAN, which decides how much part of the residual item will affect the generated image. Except for giving a fixed number, we can also set the $k$ as a learnable parameter and update $k$ in every iteration. The experimental results in Table \ref{tab:ablation} show that the fixed $k=1$ reaches a better result. 

\subsection{Output Space Adaptation}
An output space adaptation is adopted to further improve the performance of ResiDualGAN. The OSA is proven to be a more effective way than feature space when facing RS images\cite{ji_generative_2021}. In this paper, the OSA improves the mIoU by 5.24\% from PotsdamIRRG to Vaihingen, 2.57\% from PotsdamRGB to Vaihingen and 1.61\% from PotsdamIRRG to BC403. The OSA can also be replaced with other methods, like self-training, to reach higher accuracy performance in the future works. A more thorough discussion of Stage B is beyond the scope of this paper. \par

\section{Conclusion}
Aiming to learn a semantic segmentation model for RS images from an annotated dataset to an unannotated dataset, ResiDualGAN has been proposed in this paper to minimize the domain gap at the pixel level. Considering the scale discrepancy of scale-invariant objects, an in-network resizer module is used, which greatly increases the segmentation accuracy of scale-invariant classes. Considering the feature of real-to-real translation of RS images, a simple but effective residual connection is utilized, which not only stabilizes the training procedure of GANs model but also improves the accuracy of results when putting together with the resizer module. Combined with an output space adaptation, we reach the state-of-the-art accuracy performance on the benchmarks, which shows the superiority and reliability of the proposed method. \par
ResiDualGAN is a simple, but stable and effective method to train an adversarial generative model for RS images cross-domain semantic segmentation tasks. However, ResiDualGAN only minimizes the pixel-level domain gap. How to combine ResiDualGAN with some adversarial discriminative methods which minimize the feature-level and output-level domain gap and self-training strategies better for higher performance in cross-domain RS images semantic segmentation is a potential topic for future works.\par

\ifCLASSOPTIONcaptionsoff
  \newpage
\fi



%
{
\scriptsize
\bibliographystyle{IEEEtran}
\begin{spacing}{0.8}           
\bibliography{uda_new}
\end{spacing}
}
%








\end{document}